\documentclass[10pt,twocolumn,letterpaper]{article}
    
    \usepackage{cvpr}
    \usepackage{times}
    \usepackage{epsfig}
    \usepackage{graphicx}
    \usepackage{amsmath}
    \usepackage{amssymb}
    \usepackage{stmaryrd}
    \usepackage{eucal}
    \usepackage{bm}
    \usepackage{epsfig}
    \usepackage{subfigure}
    \usepackage{caption}
    \usepackage{xspace}
    \usepackage{enumerate}
    \usepackage{multirow}
    \usepackage[pagebackref=true,breaklinks=true,letterpaper=true,colorlinks,bookmarks=false]{hyperref}
    
    \usepackage{authblk}
    
    \cvprfinalcopy

    \def\cvprPaperID{1720} 

\begin{document}
    
    \pagenumbering{gobble}
    \thispagestyle{empty}

    \title{Learning Compact Recurrent Neural Networks\\ with Block-Term Tensor Decomposition}

    \vspace{-1ex}
    \author[1]{Jinmian Ye}
    \author[2]{Linnan Wang}
    \author[1]{Guangxi Li}
    \author[1]{Di Chen}
    \author[3]{Shandian Zhe}
    \author[4]{Xinqi Chu}
    \author[1]{Zenglin Xu}
    \affil[1]{{\tt \{jinmian.y,gxli2017,chendi1995425,zenglin\}@gmail.com}, SMILE Lab, University of Electronic Science and Technology of China, Chengdu, Sichuan, China}
    \affil[2]{{\tt wangnan318@gmail.com}, Dept. of Computer Science, Brown University, Providence, RI, USA}
    \affil[3]{{\tt zhe@cs.utah.edu}, School of Computing, University of Utah, Salt Lake City, Utah, USA}
    \affil[4]{{\tt ethan@xjeralabs.com}, Xjera Labs, Pte.Ltd, Singapore}

    \renewcommand\Authands{ and }
    
    
    \date{}
    
    \maketitle
    
    \begin{abstract}
Recurrent Neural Networks (RNNs) are powerful sequence modeling tools. However, when dealing with high dimensional inputs, the training of RNNs becomes computational expensive due to the large number of model parameters. This hinders RNNs from solving many important computer vision tasks, such as Action Recognition in Videos and Image Captioning. To overcome this problem, we propose a compact and flexible structure, namely Block-Term tensor decomposition, which greatly reduces the parameters of RNNs and improves their training efficiency. Compared with alternative low-rank approximations, such as tensor-train RNN (TT-RNN), our method, Block-Term RNN (BT-RNN), is not only more concise (when using the same rank), but also able to attain a better approximation to the original RNNs with much fewer parameters.  On three challenging tasks, including Action Recognition in Videos, Image Captioning and Image Generation, BT-RNN outperforms TT-RNN and the standard RNN in terms of both prediction accuracy and convergence rate. Specifically, BT-LSTM utilizes 17,388 times fewer parameters than the standard LSTM to achieve an accuracy improvement over 15.6\% in the Action Recognition task on the UCF11 dataset. 
\end{abstract}

    \section{Introduction}








Best known for the sequence-to-sequence learning, the Recurrent Neural Networks (RNNs) belong to a class of neural architectures designed to capture the dynamic temporal behaviors of data. The vanilla fully connected RNN utilizes a feedback loop to memorize previous information, while it is inept to handle long sequences as the gradient exponentially vanishes along the time \cite{hochreiter1991untersuchungen, bengio1994learning}. Unlike the vanilla RNNs passing information between layers with direct matrix-vector multiplications, the Long Short-Term Memory (LSTM) introduces a number of gates and passes information with element-wise operations~\cite{hochreiter1997long}. This improvement drastically alleviates the gradient vanishing issue; 
therefore LSTM and its variants, e.g. Gated Recurrent Unit (GRU) \cite{cho2014learning}, are widely used in various Computer Vision (CV) tasks \cite{byeon2015scene, liang2016semantic, theis2015generative} to model the long-term correlations in sequences. 

\begin{figure}[t]
	\centering
	\includegraphics[width=0.45\textwidth]{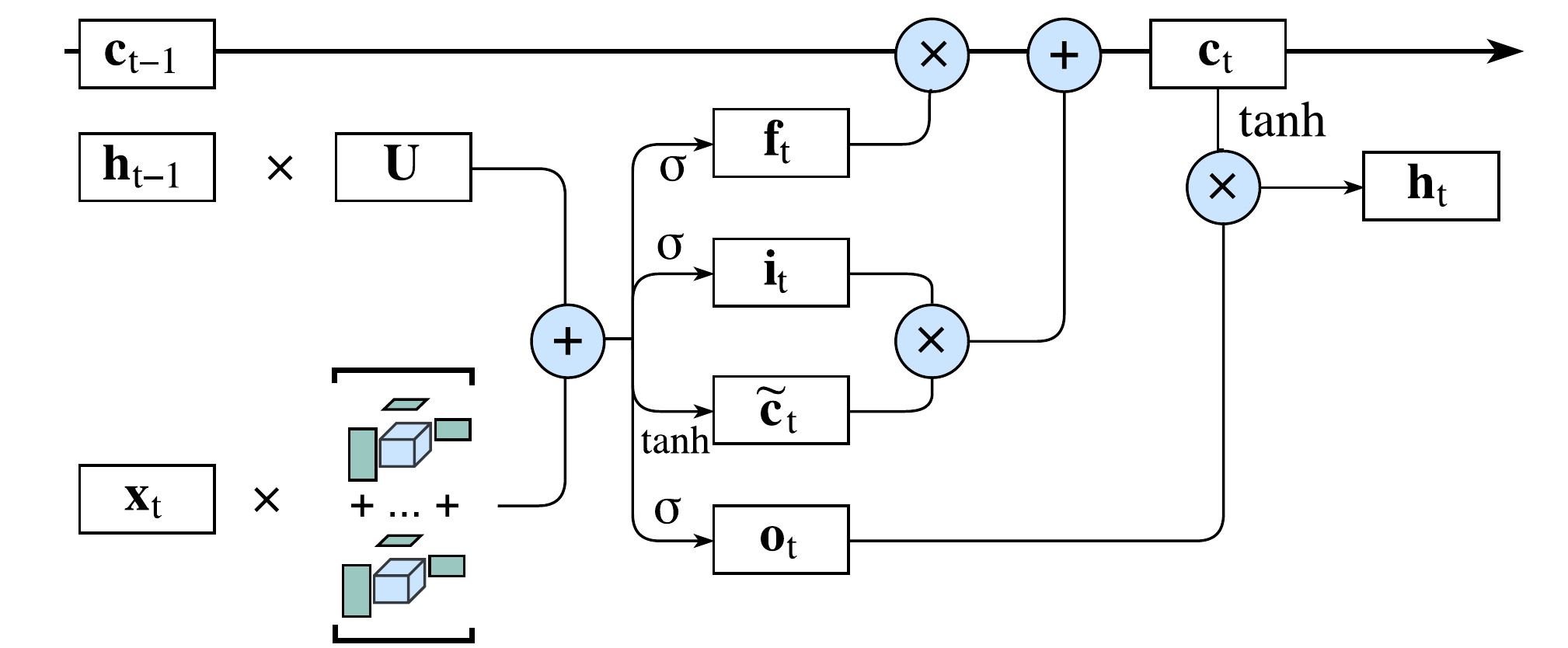}
	\caption{Architecture of BT-LSTM. The redundant dense connections between input and hidden state is replaced by low-rank BT representation.}
	\vspace{-2ex}
	\label{fig:architecture}
\end{figure}

The current formulation of LSTM, however, suffers from an excess of parameters, making it notoriously difficult to train and susceptible to overfitting. 
The formulation of LSTM can be described by the following equations:
\begin{align}
	\mathbf{f}_t &= \sigma(\mathbf{W}_f \cdot \mathbf{x}_t + \mathbf{U}_f \cdot \mathbf{h}_{t-1}+ \mathbf{b}_f ) \\
	\mathbf{i}_t &= \sigma(\mathbf{W}_i \cdot \mathbf{x}_t + \mathbf{U}_i \cdot \mathbf{h}_{t-1} + \mathbf{b}_i ) \\
	\mathbf{o}_t &= \sigma(\mathbf{W}_o \cdot \mathbf{x}_t + \mathbf{U}_{o} \cdot \mathbf{h}_{t-1} + \mathbf{b}_o ) \\
	\tilde{\mathbf{c}}{_t} &= \tanh(\mathbf{W}_c \cdot \mathbf{x}_t + \mathbf{U}_{c} \cdot \mathbf{h}_{t-1} + \mathbf{b}_c) \\
	\mathbf{c}_t &= \mathbf{f}_t \odot \mathbf{c}_{t-1} + \mathbf{i}_t \odot \tilde{\mathbf{c}_t} \\
	\mathbf{h}_t &= \mathbf{o}_t \odot \tanh(\mathbf{c}_t),
\end{align}
where $\odot$ denotes the element-wise product, $\sigma(\cdot)$ denotes the sigmoid function and $\tanh(\cdot)$ is the hyperbolic tangent function. The weight matrices $\mathbf{W}_*$ and $\mathbf{U}_*$ transform the input $\mathbf{x}_t$ and the hidden state $\mathbf{h}_{t-1}$, respectively, to cell update $\tilde{\mathbf{c}}{_t}$ and three gates $\mathbf{f}_t$, $\mathbf{i}_t$, and $\mathbf{o}_t$.
Please note that given an image feature vector $\mathbf{x}_t$ fetch from a Convolutional Neural Network (CNN) network, the shape of $\mathbf{x}_t$ will raise to $I = 4096$ and $I = 14 \times 14 \times 512$ w.r.t vgg16 \cite{simonyan2014very} and Inception v4 \cite{szegedy2017inception}. If the number of hidden states is $J=256$, the total number of parameters in calculating the four $\mathbf{W}_*$ is $4 \times I \times J$, which can up to $4.1 \times 10^6$ and $1.0 \times 10^8$, respectively. Therefore, the giant matrix-vector multiplication, i.e., $\mathbf{W}_* \cdot \mathbf{x}_t$, leads to the major inefficiency -- the current parameter-intensive design not only subjects the model difficult to train, but also lead to high computation complexity and memory usage. 


In addition, each $\mathbf{W}_* \cdot \mathbf{x}_t $ essentially represents a fully connected operation that transforms the input vector $\mathbf{x}_t$ into the hidden state vector. However, extensive research on CNNs has proven that the dense connection is significantly inefficient at extracting the spatially latent local structures and local correlations naturally exhibited in the image~\cite{lecun1995convolutional, han2015learning}. 
Recent leading CNNs architectures, e.g., DenseNet \cite{huang2016densely}, ResNet \cite{he2016deep} and Inception v4 \cite{szegedy2017inception}, also try to circumvent one huge cumbersome dense layer \cite{szegedy2015going}. 
But the discussions of improving the dense connections in RNNs are still quite limited
\cite{lu2016learning, prabhavalkar2016compression}.
It is imperative to seek a more efficient design to replace $\mathbf{W}_* \cdot \mathbf{x}_t$.


In this work, we propose to design a sparsely connected tensor representation, i.e., the Block-Term decomposition (BTD) \cite{de2008decompositions2}, to replace the redundant and densely connected operation in LSTM \footnote{we focus on LSTM in this paper, but the proposed approach also applies for other variants such as GRU.}. The Block-Term decomposition is a low-rank approximation method that decomposes a high-order tensor into a sum of multiple Tucker decomposition models~\cite{tucker1966some,XuYQ12icmltensor,XuYQ15tensor,liyycyx17tensor}. 
In detail, we represent the four weight matrices (i.e.,  $\mathbf{W}_*$) and the input data $\mathbf{x}_t$ into a various order of tensor.
In  the  process of RNNs training, the BTD layer automatically learns inter-parameter correlations to implicitly prune redundant dense connections rendered by $\mathbf{W}\cdot \mathbf{x}$. By plugging the new BTD layer into current RNNs formulations, we present a new BT-RNN model with a similar representation power but several orders of fewer parameters. The refined LSTM model with the Block-term representation is illustrated in  Fig. \ref{fig:architecture}.






The major merits of BT-RNN are shown as follows:
\begin{itemize}
\item The low-rank BTD can compress the dense connections in the input-to-hidden transformation, while still retaining the current design philosophy of LSTM. By reducing several orders of model parameters, BT-LSTM has better convergence rate than the traditional LSTM architecture, significantly enhancing the training speed.
\item  Each dimension in the input data can share weights with all the other dimensions as the existence of core tensors, thus BT representation has the strong connection between different dimensions, enhancing the ability to capture sufficient local correlations. Empirical results show that, compared with the Tensor Train model~\cite{oseledets2011tensor}, the BT model has a better representation power with the same amount of  model parameters.
\item The design of multiple Tucker models can significantly reduce the sensitivity to noisy input data and widen network, leading to a more robust RNN model. In contrary to the Tensor Train based tensor approaches \cite{yang2017tensor, novikov2015tensorizing}, the BT model does not suffer from the difficulty of ranks setting, releasing researchers from intolerable work in choosing hyper-parameters.
\end{itemize}

In order to demonstrate the performance of the BT-LSTM model, we design three challenging computer vision tasks -- Action Recognition in Videos, Image Caption and Image Generation -- to quantitatively and qualitatively evaluate the proposed BT-LSTM against the baseline LSTM and other low-rank variants such as the Tensor Train LSTM (TT-LSTM). Experimental results have demonstrated the promising performance of the BT-LSTM model.







    \section{Related Work}


The poor image modeling efficiency of full connections in the perception architecture, i.e., $\mathbf{W} \cdot \mathbf{x}$ \cite{wang2016blasx}, has been widely recognized by the Computer Vision (CV) community. The most prominent example is the great success made by Convolutional Neural Networks (CNNs) for the general image recognition. Instead of using the dense connections in multi-layer perceptions, CNNs relies on sparsely connected convolutional kernels to extract the latent regional features in an image. Hence, going sparse on connections is the key to the success of CNNs \cite{denil2013predicting, jaderberg2014speeding, molchanov2016pruning, hinton2015distilling, srivastava2014dropout}. Though extensive discussions toward the efficient CNNs design, the discussions of improving the dense connections in RNNs are still quite limited
\cite{lu2016learning, prabhavalkar2016compression}.


Compared with aforementioned explicit structure changes, the low-rank method is one orthogonal approach to implicitly prune the dense connections.
Low-rank tensor methods have been successfully applied to address the redundant dense connection problem in CNNs~\cite{novikov2015tensorizing, yang2017tensor, bai2017tensorial, tjandra2017compressing, kossaifi2017tensor}.
Since the key operation in one perception is $\mathbf{W} \cdot \mathbf{x}$, Sainath et al. \cite{sainath2013low} decompose $\mathbf{W}$ with Singular Value Decomposition (SVD), reducing up to 30\% parameters in $\mathbf{W}$, but also demonstrates up to 10\% accuracy loss \cite{xue2013restructuring}. The accuracy loss majorly results from losing the high-order spatial information, as intermediate data after image convolutions are intrinsically in 4D.

In order to capture the high order spatial correlations, recently, tensor methods were introduced into Neural Networks to approximate $\mathbf{W} \cdot \mathbf{x}$. For example, Tensor Train (TT) method was employed to alleviate the large computation $\mathbf{W} \cdot \mathbf{x}$ and reduce the number of parameters \cite{novikov2015tensorizing, yang2017tensor, tjandra2017compressing}. 
Yu et al. \cite{yu2017long} also used a tensor train representation to forecast long-term information.
Since this approach targets in long historic states, it increases additional parameters, leading to a difficulty in training. Other tensor decomposition methods also applied in Deep Neural Networks (DNNs) for various purposes \cite{lebedev2014speeding, yunpeng2017sharing, kossaifi2017tensor}.




Although TT decomposition has obtained a great success in addressing dense connections problem, there are some limitations which block TT method to achieve better performance: 1) The optimal setting of TT-ranks is that they are small in the border cores and large in middle cores, e.g., like an olive \cite{zhao2017learning}. However, in most applications, TT-ranks are set equally, which will hinder TT's representation ability. 2) TT-ranks has a strong constraint that the rank in border tensors must set to 1 ($R_1 = R_{d+1} = 1$), leading to a seriously limited representation ability and flexibility \cite{yang2017tensor, zhao2017learning}.


Instead of difficultly finding the optimal TT-ranks setting, BTD has these advantages: 1) Tucker decomposition introduces a core tensor to represent the correlations between different dimensions, achieving better weight sharing. 2) ranks in core tensor can be set to equal, avoiding unbalance weight sharing in different dimensions, leading to a robust model toward different permutations of input data. 3) BTD uses a sum of multiple Tucker models to approximate a high-order tensor, breaking a large Tucker decomposition to several smaller models, widening network and increasing representation ability. Meanwhile, multiple Tucker models also lead to a more robust RNN model to noisy input data.

    \section{Tensorizing Recurrent Neural Networks}
The core concept of this work is to approximate $\mathbf{W} \cdot \mathbf{x}$ with much fewer parameters, while still preserving the memorization mechanism in existing RNN formulations. The technique we use for the approximation is Block Term Decomposition (BTD), which represents $\mathbf{W} \cdot \mathbf{x}$ as a series of light-weighted small tensor products. In the process of RNN training, the BTD layer automatically learns inter-parameter correlations to implicitly prune redundant dense connections rendered by $\mathbf{W} \cdot \mathbf{x}$. By plugging the new BTD layer into current RNN formulations, we present a new BT-RNN model with several orders of magnitude fewer parameters while maintaining the representation power.

This section elaborates the details of the proposed methodology. It starts with exploring the background of tensor representations and BTD, before delving into the transformation of a regular RNN model to the BT-RNN; then we present the back propagation procedures for the BT-RNN; finally, we analyze the time and memory complexity of the BT-RNN compared with the regular one.

\subsection{Preliminaries and Background}
\paragraph{Tensor Representation} We use the boldface Euler script letter, e.g., $\bm{\mathcal{X}}$, to denote a tensor. A $d$-order tensor represents a $d$ dimensional multiway array; thereby a vector and a matrix is a 1-order tensor and a 2-order tensor, respectively.
An element in a $d$-order tensor is denoted as $\bm{\mathcal{X}}_{i_1,\dots,i_d}$. 

\vspace{-1ex}
\paragraph{Tensor Product and Contraction} Two tensors can perform product on a $kth$ order if their $kth$ dimension matches. 
Let's denote $\bullet_k$ as the tensor-tensor product on $kth$ order \cite{kolda2009tensor}. Given two $d$-order tensor $\bm{\mathcal{A}} \in \mathbb{R}^{I_1 \times \dots \times I_d}$ and $\bm{\mathcal{B}} \in \mathbb{R}^{J_1 \times \dots \times J_d}$, the tensor product on $kth$ order is:
\begin{align}
    (\bm{\mathcal{A}} \bullet_k \bm{\mathcal{B}})_{\bm{i_k^-},\bm{i_k^+},\bm{j_k^-},\bm{j_k^+}} = 
    \sum_{p=1}^{I_k} \bm{\mathcal{A}}_{\bm{i_k^-}, p, \bm{i_k^+}} \bm{\mathcal{B}}_{\bm{j_k^-}, p, \bm{j_k^+}}.
\end{align}
To simplify, we use $\bm{i_k^-}$ denotes indices $(i_1,\dots,i_{k-1})$, while $\bm{i_k^+}$ denotes $(i_{k+1},\dots,i_d)$. The whole indices can be denoted as $\vec{\bm{i}} := (\bm{i_k^-},i_k,\bm{i_k^+})$. As we can see that each tensor product will be calculated along $I_k$ dimension, which is consistent with matrix product.

Contraction is an extension of tensor product \cite{cichocki2014era}; it conducts tensor products on multiple orders at the same time. For example, if $I_k=J_k, I_{k+1}=J_{k+1}$, we can conduct a tensor product according the $kth$ and $(k+1)th$ order:
\begin{align}
    (\bm{\mathcal{A}} \bullet_{k, k+1} \bm{\mathcal{B}})&_{\bm{i_k^-},\bm{i_{k+1}^+}, \bm{j_k^-}, \bm{j_{k+1}^+}} = \nonumber \\
    &\sum_{p=1}^{I_k} \sum_{q=1}^{I_{k+1}} \bm{\mathcal{A}}_{\bm{i_k^-},p,q, \bm{i_{k+1}^+}} \bm{\mathcal{B}}_{\bm{j_k^-},p,q, \bm{j_{k+1}^+}}.
\end{align}

\begin{figure}[t]
    \centering
    \includegraphics[width=0.45\textwidth]{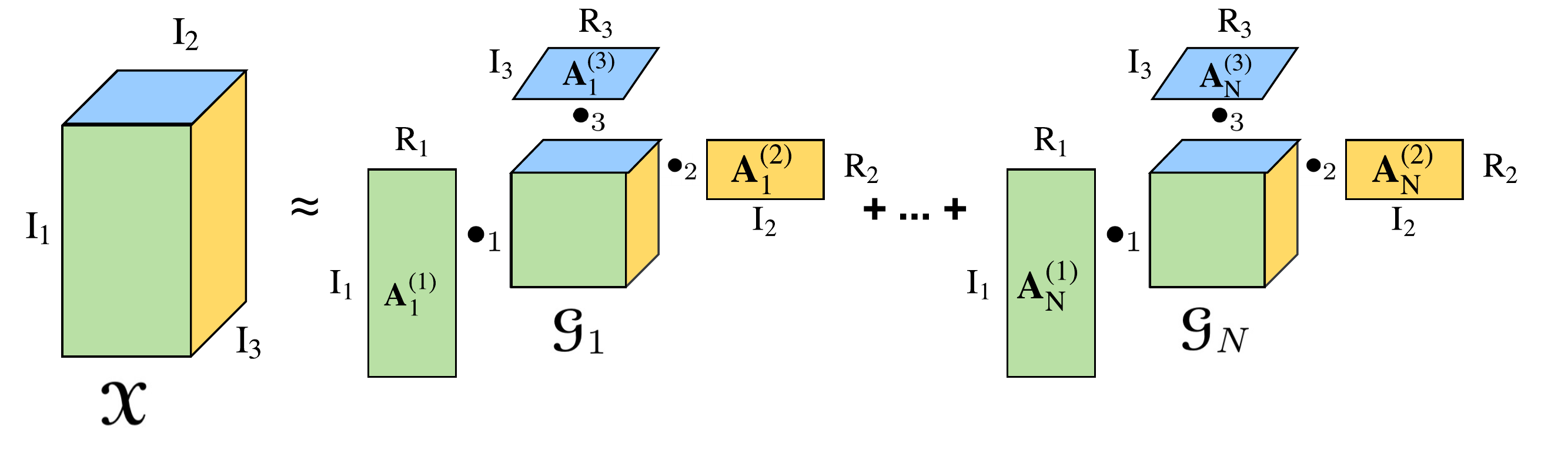}
    \caption{Block Term decomposition for a 3-order case tensor. A 3-order tensor $\bm{\mathcal{X}} \in \mathbb{R}^{I_1 \times I_2 \times I_3}$ can be approximated by $N$ Tucker decompositions. We call the $N$ the CP-rank, $R_1,R_2,R_3$ the Tucker-rank and $d$ the Core-order.}
    \label{fig:bt-decomp}
    \vspace{-2ex}
\end{figure}

\paragraph{Block Term Decomposition (BTD)}
Block Term decomposition is a combination of CP decomposition \cite{carroll1970analysis} and Tucker decomposition \cite{tucker1966some}.
Given a $d$-order tensor $\bm{\mathcal{X}} \in \mathbb{R}^{I_1 \times \dots \times I_d } $, BTD decomposes it into $N$ block terms; And each term conducts $\bullet_k$ between a core tensor $\bm{\mathcal{G}}_n \in \mathbb{R}^{R_1 \times \dots \times R_d} $ and $d$ factor matrices $\bm{\mathcal{A}}_n^{(k)} \in \mathbb{R}^{I_k \times R_k }$ on $\bm{\mathcal{G}}_n$'s $kth$ dimension, where $n \in [1, N]$ and $k \in [1, d]$ \cite{de2008decompositions2}. The formulation of BTD is as follows:
\begin{align}
    \label{BTD_formulation}
    \bm{\mathcal{X}} = \sum_{n=1}^N \bm{\mathcal{G}}_n \bullet_1  \bm{\mathcal{A}}_n^{(1)} \bullet_2 \bm{\mathcal{A}}_n^{(2)} \bullet_3 \dots \bullet_d \bm{\mathcal{A}}_n^{(d)}.
\end{align}
We call the $N$ the CP-rank, $R_1,R_2,R_3$ the Tucker-rank and $d$ the Core-order.
Fig. \ref{fig:bt-decomp} demonstrates an example of how 3-order tensor $\bm{\mathcal{X}}$ being decomposed into $N$ block terms.

\subsection{BT-RNN model}
This section demonstrates the core steps of BT-RNN model. 1) We transform $\mathbf{W}$ and $\mathbf{x}$ into tensor representations, $\bm{\mathcal{ W }}$ and $\bm{\mathcal{ X }}$;  2) then we decompose $\bm{\mathcal{ W }}$ into several low-rank core tensors $\bm{\mathcal{G}}_n$ and their corresponding factor tensors $\bm{\mathcal{A}}_n^{(d)}$ using BTD; 3) subsequently, the original product $\mathbf{W} \cdot \mathbf{x}$ is approximated by the tensor contraction between decomposed weight tensor $\bm{\mathcal{ W }}$ and input tensor $\bm{\mathcal{ X }}$; 4) finally, we present the gradient calculations amid Back Propagation Through Time (BPTT) \cite{werbos1990backpropagation, wang2017accelerating} to demonstrate the learning procedures of BT-RNN model. 


\begin{figure}[t]
    \centering
    \subfigure[vector to tensor]{\label{fig:vector2tensor}
    \includegraphics[width=0.23\textwidth]{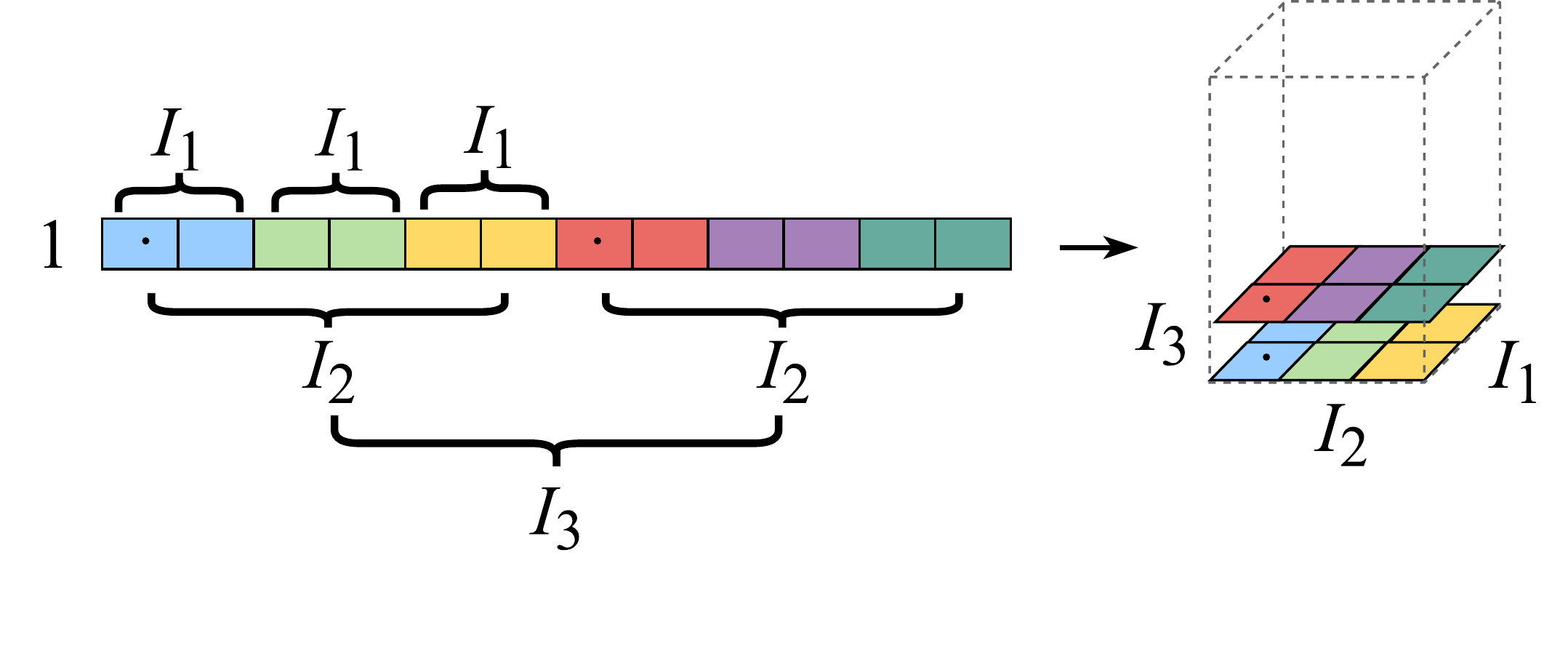}}
    \subfigure[matrix to tensor]{\label{fig:matrix2tensor}
    \includegraphics[width=0.23\textwidth]{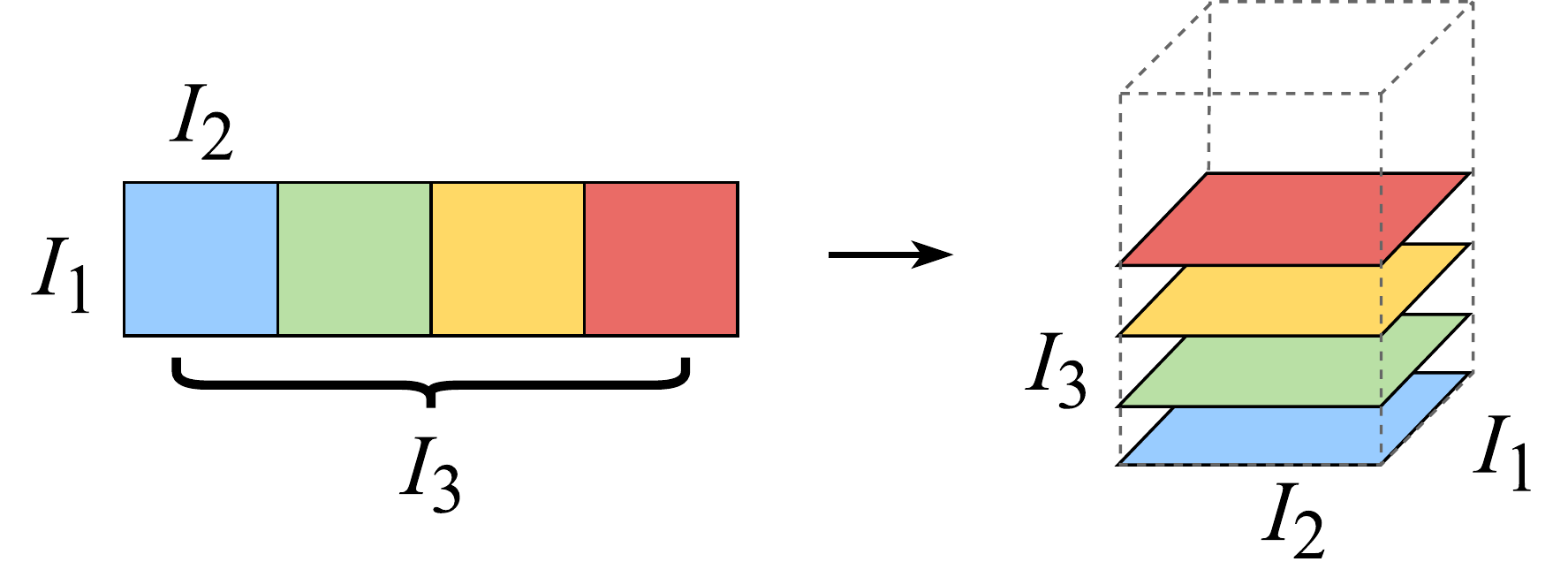}}
    \caption{Tensorization operation in a case of 3-order tensors. (a) Tensorizing a vector with shape $I = I_1 \cdot  I_2 \cdot I_3$ to a tensor with shape $I_1 \times I_2 \times I_3$; (b) Tensorizing a matrix with shape $I_1 \times (I_2 \cdot I_3)$ to a tensor with shape $I_1 \times I_2 \times I_3$.}
    \label{fig:vector_matrix_tensor}
    \vspace{-2ex}
\end{figure}

\vspace{-1ex}
\paragraph{Tensorizing $\mathbf{W}$ and $\mathbf{x}$} we \emph{tensorize} the input vector $\mathbf{x}$ to a high-order tensor $\bm{\mathcal{ X }}$ to capture spatial information of the input data, while we \emph{tensorize} the weight matrix $\mathbf{W}$ to decomposed weight tensor $\bm{\mathcal{ W }}$ with BTD. 

Formally, given an input vector $\mathbf{x}_t \in \mathbb{R}^{I} $, we define the notation $\varphi$ to denote the \emph{tensorization} operation. It can be either a \emph{stack} operation or a \emph{reshape} operation. We use reshape operation for tensorization as it does not need to duplicate the element of the data. Essentially reshaping is regrouping the data. Fig. \ref{fig:vector_matrix_tensor} outlines how we reshape  a vector and a matrix into  3-order tensors.

\vspace{-1ex}
\paragraph{Decomposing $\mathbf{W}$ with BTD} Given a 2 dimensions weight matrix $\mathbf{W} \in \mathbb{R}^{J \times I}$, we can tensorize it as a $2d$ dimensions tensor $\bm{\mathcal{W}} \in \mathbb{R}^{J_1 \times I_1 \times J_2 \times\dots \times J_d \times I_d}$, where $I = I_1 I_2  \cdots  I_d $ and $J = J_1  J_2  \cdots J_d$. 
Following BTD in Eq. (\ref{BTD_formulation}), we can decomposes $\bm{\mathcal{W}}$ into:
\begin{align}
    BTD(\bm{\mathcal{W}}) = \sum_{n=1}^N \bm{\mathcal{G}}_n \bullet_1 \bm{\mathcal{A}}_n^{(1)} \bullet_2 \cdots \bullet_d \bm{\mathcal{A}}_n^{(d)},
    \label{eqt:bt_model}
    \vspace{-2ex}
\end{align}
where $\bm{\mathcal{G}}_n \in \mathbb{R}^{R_1 \times \dots \times R_d}$ denotes the core tensor, $\bm{\mathcal{A}}_n^{(d)} \in \mathbb{R}^{I_d \times J_d \times R_d}$ denotes the factor tensor, $N$ is the CP-rank and $d$ is the Core-order. From the mathematical property of BT's ranks \cite{kolda2009tensor}, we have $R_k \leqslant I_k$ (and $J_k$), $k=1,\ldots, d$. If $R_k > I_k$ (or $J_k$), it is difficult for the model to obtain bonus in performance. What's more, to obtain a robust model, in practice, we set each Tucker-rank to be equal, e.g., $R_i = R$, $i \in [1, d]$, to avoid unbalanced weight sharing in different dimensions and to alleviate the difficulty in hyper-parameters setting.

\begin{figure}[t]
    \centering
    \includegraphics[width=0.40\textwidth]{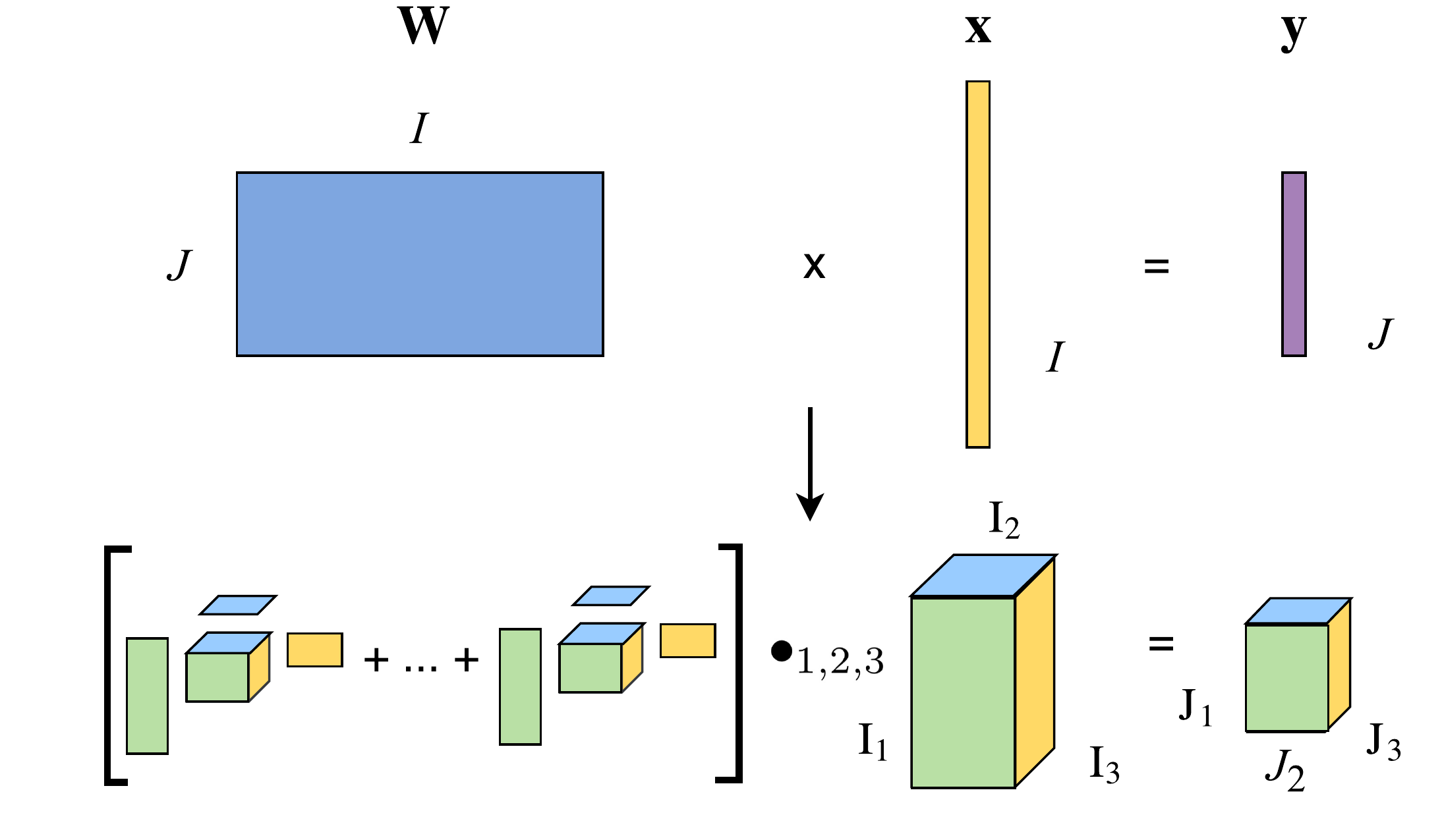}
    \caption{Diagrams of BT representation for matrix-vector product $\mathbf{y} = \mathbf{Wx}, \mathbf{W} \in \mathbb{R}^{J \times I}$. We substitute the weight matrix $\mathbf{W}$ by the BT representation, then tensorize the input vector $\mathbf{x}$ to a tensor with shape $I_1 \times I_2 \times I_3$. After operating the tensor contraction between BT representation and input tensor, we get the result tensor in shape $J_1 \times J_2 \times J_3$. With the reverse tensorize operation, we get the output vector $\mathbf{y} \in \mathbb{R}^{J_1 \cdot J_2 \cdot J_3}$. }
    \label{fig:tensorproduct}
    \vspace{-2ex}
\end{figure}

\vspace{-1ex}
\paragraph{Computation between $\mathbf{W}$ and $\mathbf{x}$} After substituting the matrix-vector product by BT representation and tensorized input vector, we replace the input-to-hidden matrix-vector product $\mathbf{W} \cdot \mathbf{x_t}$ with the following form:
\begin{align}
    \phi(\mathbf{W}, \mathbf{x}_t) = BTD(\bm{\mathcal{W}}) \bullet_{1,2,\ldots,d} \bm{\mathcal{X}}_t,
    \label{eqt:bt_representation}
\end{align}
where the tensor contraction operation $\bullet_{1,2,\ldots,d}$ will be computed along all $I_k$ dimensions in $\bm{\mathcal{W}}$ and $\bm{\mathcal{X}}$, yielding the same size in the element-wise form as the original one. Fig. \ref{fig:tensorproduct} demonstrates the substitution intuitively.

\vspace{-1ex}
\paragraph{Training BT-RNN} The gradient of RNN is computed by Back Propagation Through Time (BPTT) \cite{werbos1990backpropagation}. We derive the gradients amid the framework of BPTT for the proposed BT-RNN model.

Following the regular LSTM back-propagation procedure, the gradient $\frac{\partial L}{\partial \mathbf{y}}$ can be computed by the original BPTT algorithm, where $\mathbf{y} = \mathbf{W} \mathbf{x}_t$. Using the \emph{tensorization} operation same to $\mathbf{y}$, we can obtain the tensorized gradient $\frac{\partial L}{\partial \bm{\mathcal{Y}}}$. For a more intuitive understanding, we rewrite Eq. (\ref{eqt:bt_representation}) in element-wise case:
\begin{align}
    \bm{\mathcal{Y}}_{\vec{\bm{j}}} = \sum_{n=1}^N & \sum_{\vec{\bm{r}}}^{R_1,\dots,R_d} \sum_{\vec{\bm{i}}}^{I_1, \ldots, I_d} \prod_{k=1}^d
    \bm{\mathcal{X}}_{t,\vec{\bm{i}}} \bm{\mathcal{A}}^{(k)}_{n,i_k, j_k, r_k} \bm{\mathcal{G}}_{n, \vec{\bm{r}}}. \label{eqt:element_wise_bt}
\end{align}
Here, for simplified writing, we use $\vec{\bm{i}}$, $\vec{\bm{j}}$ and $\vec{\bm{r}}$ to denote the indices $(i_1, \dots, i_d)$, $(j_1, \dots, j_d)$ and $(r_1, \dots, r_d)$, respectively.
Since the right hand side of Eq. (\ref{eqt:element_wise_bt}) is a scalar, the element-wise gradient for parameters in BT-RNN is as follows:
\begin{align}
    &\frac{\partial L}{\partial \bm{\mathcal{A}}_{n, i_k,j_k,r_k}^{(k)}} = \nonumber \\
     &\sum_{\vec{\bm{r}} \neq r_k } 
     \sum_{\vec{\bm{i}} \neq i_k } 
     \sum_{\vec{\bm{j}} \neq j_k }
     \prod_{{k'} \neq k}^d
     \bm{\mathcal{G}}_{n, \vec{\bm{r}}}
     \bm{\mathcal{A}}_{n, i_{k'}, j_{k'}, r_{k'}}  
     \bm{\mathcal{X}}_{t,\vec{\bm{i}}} \frac{\partial L}{\partial \bm{\mathcal{Y}}_{\vec{\bm{j}}}},  \\
    &\frac{\partial L}{\partial \mathcal{G}_{n, \vec{\bm{r}}}} = 
    \sum_{\vec{\bm{i}}}
    \sum_{\vec{\bm{j}}}
    \prod_{k=1}^d \bm{\mathcal{A}}_{n, i_k, j_k, r_k}
    \bm{\mathcal{X}}_{t,\vec{\bm{i}}}
    \frac{\partial L}{\partial \bm{\mathcal{Y}}_{\vec{\bm{j}}}}.
\end{align}

\subsection{Hyper-Parameters and Complexity Analysis}
\subsubsection{ Hyper-Parameters Analysis }
\paragraph{Total \#Params}
BTD decomposes $ \bm{\mathcal{W}} $ into $N$ block terms and each block term is a tucker representation \cite{kossaifi2017tensor, kolda2009tensor}, therefore the total amount of parameters is as follows:
\begin{align}
    P_{BTD} = N (\sum_{k=1}^d I_k J_k R + R^d).
    \vspace{-2ex}
    \label{eqt:number_of_parameters}
\end{align}
By comparison, the original weight matrix $\mathbf{W}$ contains $P_{RNN} = I\times J = \prod_{k=1}^d I_k J_k$ parameters, which is several orders of magnitude larger than it in the BTD representation.

\begin{figure}[t]
    \centering
    \includegraphics[width=0.45\textwidth]{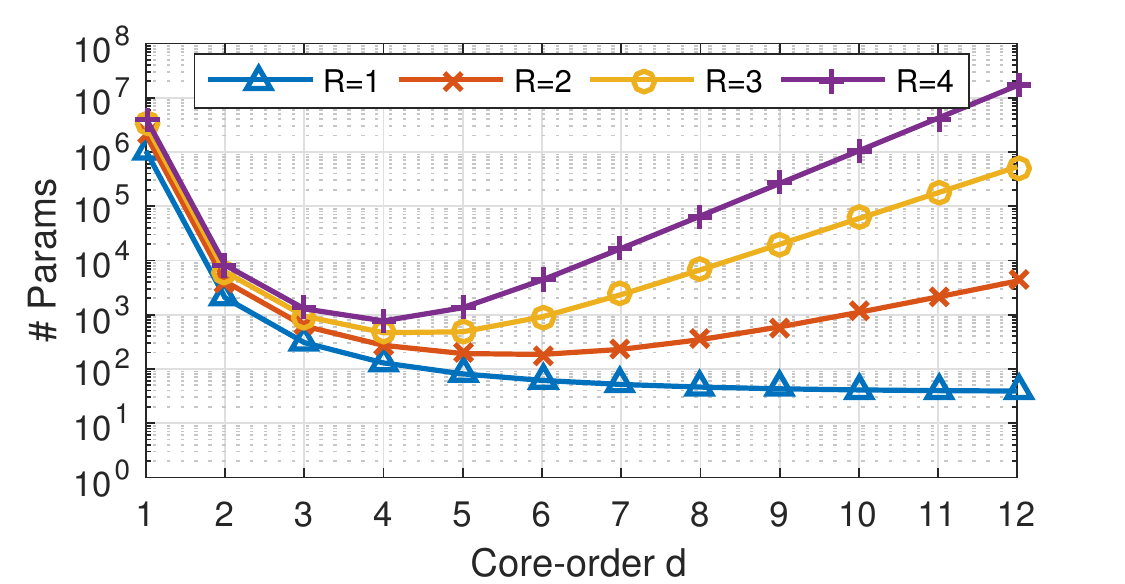}
    \caption{The number of parameters w.r.t Core-order $d$ and Tucker-rank $R$, in the setting of $I = 4096, J = 256, N = 1$. While the vanilla RNN contains $I\times J = 1048576$ parameters. Refer to Eq. (\ref{eqt:number_of_parameters}), when $d$ is small, the first part $\sum_1^d I_k J_k R$ does the main contribution to parameters. While $d$ is large, the second part $R^d$ does. So we can see the number of parameters will go down sharply at first, but rise up gradually as $d$ grows up (except for the case of $R=1$).}
    \label{fig:paramd}
    \vspace{-2ex}
\end{figure}

\vspace{-1ex}
\paragraph{\#Params w.r.t Core-order ($d$)} Core-order $d$ is the most significant factor affecting the total amount of parameters as $R^d$ term in Eq. (\ref{eqt:number_of_parameters}). It determines the total dimensions of core tensors, the number of factor tensors, and the total dimensions of input and output tensors. If we set $d=1$, the model degenerates to the original matrix-vector product with the largest number of parameters and the highest complexity. Fig. \ref{fig:paramd} demonstrates how total amount of parameters vary w.r.t different Core-order $d$. If the Tucker-rank $R > 1$, the total amount of parameters first decreases with $d$ increasing until reaches the minimum, then starts increasing afterwards. This mainly results from the non-linear characteristic of $d$ in Eq. (\ref{eqt:number_of_parameters}). 

Hence, a proper choice of $d$ is particularly important. Enlarging the parameter $d$ is the simplest way to reduce the number of parameters. But due to the second term $R^d$ in Eq. (\ref{eqt:number_of_parameters}), enlarging $d$ will also increase the amount of parameters in the core tensors, resulting in the high computational complexity and memory usage. With the Core-order $d$ increasing, each dimensions of the input tensor decreases logarithmically. However, this will result in the loss of important spatial information in an extremely high order BT model. In practice, Core-order $d \in [2, 5]$ is recommended.

\vspace{-1ex}
\paragraph{\#Params w.r.t Tucker-rank ($R$) }
The Tucker-rank $R$ controls the complexity of Tucker decomposition. This hyper-parameter is conceptually similar to the number of singular values in Singular Value Decomposition (SVD). Eq. (\ref{eqt:number_of_parameters}) and Fig. \ref{fig:paramd} also suggest the total amount of parameters is sensitive to $R$. Particularly, BTD degenerates to a CP decomposition if we set it as $R = 1$. Since $R \leqslant I_k$ (and  $J_k$), the choice of $R$ is limited in a small value range, releasing researchers from heavily hyper-parameters setting.

\vspace{-1ex}
\paragraph{\#Params w.r.t CP-rank ($N$)}
The CP-rank $N$ controls the number of block terms. If $N = 1$, BTD degenerates to a Tucker decomposition. As we can see from Table \ref{tbl:complexity} that $N$ does not affect the memory usage in forward and backward passes, so if we need a more memory saving model, we can enlarge $N$ while decreasing $R$ and $d$ at the same time.

\subsubsection{Computational Complexity Analysis}
\paragraph{Complexity in Forward Process}
Eq. (\ref{eqt:bt_model}) raises the computation peak, $\mathcal{O}(IJR)$, at the last tensor product $\bullet_d$, according to left-to-right computational order. However, we can reorder the computations to further reduce the total model complexity $\mathcal{O}(NdIJR)$. The reordering is:
\begin{align}
    \phi(\mathbf{W}, \mathbf{x}_t) = \sum_{n=1}^N \bm{\mathcal{X}}_t \bullet_1 \bm{\mathcal{A}}_n^{(1)} \bullet_2 \dots \bullet_d \bm{\mathcal{A}}_n^{(d)} \bullet_{1,2,\ldots,d} \bm{\mathcal{G}}_n. \label{eqt:bt_reorder}
\end{align}
The main difference is each tensor product will be first computed along all $R$ dimensions in Eq. (\ref{eqt:bt_representation}), while in Eq. (\ref{eqt:bt_reorder}) along all $I_k$ dimensions. Since BTD is a low-rank decomposition method, e.g., $R \leqslant J_k$ and $J_{max} R^{(d-1)} \leqslant J$, the new computation order can significantly reduce the complexity of the last tensor product from $\mathcal{O}(I J R)$ to $\mathcal{O}(I J_{max} R^d)$, where $J=J_1J_2\cdots J_d$, $J_{max} = \max_k(J_k)$, $k \in [1, d]$.
And then the total complexity of our model reduces from $\mathcal{O}(Nd I J R)$ to $\mathcal{O}(Nd I J_{max} R^d)$.
If we decrease Tucker-rank $R$, the computation complexity decreases logarithmically in Eq. (\ref{eqt:bt_reorder}) while linearly in Eq. (\ref{eqt:bt_representation}).

\vspace{-1ex}
\paragraph{Complexity in Backward Process}
To derive the computational complexity in the backward process, we present gradients in the tensor product form. The gradients of factor tensors and core tensors are:
\begin{align}
    \frac{\partial L}{\partial \bm{\mathcal{A}}_n^{(k)}} =&
    \frac{\partial L}{\partial \bm{\mathcal{Y}}}
    \bullet_k \bm{\mathcal{X}}_t \bullet_1 \bm{\mathcal{A}}_n^{(1)} \bullet_2 \dots \label{eqt:partial_A}\\
    &\bullet_{k-1} \bm{\mathcal{A}}_n^{(k-1)} \bullet_{k+1} \bm{\mathcal{A}}_n^{(k+1)} \bullet_{k+2} \dots \nonumber \\
    &\bullet_d \bm{\mathcal{A}}_n^{(d)} \bullet_{1,2,\dots,d} \bm{\mathcal{G}}_n \nonumber \\
    \frac{\partial L}{\partial \bm{\mathcal{G}}_n} =& \bm{\mathcal{X}}_t \bullet_1 \bm{\mathcal{A}}_n^{(1)} \bullet_2 \dots \bullet_d \bm{\mathcal{A}}_n^{(d)} \bullet_{1,2,\dots,d} \frac{\partial L}{\partial \bm{\mathcal{Y}}} \label{eqt:partial_G}
\end{align}
Since Eq. (\ref{eqt:partial_A}) and Eq. (\ref{eqt:partial_G}) follow the same form of Eq. (\ref{eqt:bt_representation}),
the backward computational complexity is same as the forward pass  $\mathcal{O}(d I J_{max} R^d)$. Therefore, the $N \cdot d$ factor tensors demonstrate a total complexity of $\mathcal{O}(N d^2 I J_{max} R^d)$.

\begin{table}[t]
    \begin{center}
    \begin{tabular}{l|l|l}
    \hline
    Method & Time & Memory \\
    \hline\hline
    RNN forward     & $\mathcal{O}(IJ)$ & $\mathcal{O}(IJ)$  \\
    RNN backward     & $\mathcal{O}(IJ)$ & $\mathcal{O}(IJ)$  \\
    TT-RNN forward     & $\mathcal{O}(d I R^2 J_{max})$         & $\mathcal{O}(R I)$ \\
    TT-RNN backward & $\mathcal{O}(d^2 I R^4 J_{max})$         & $\mathcal{O}(R^3 I)$  \\
    BT-RNN forward     & $\mathcal{O}(N d I R^d J_{max})$      & $\mathcal{O}(R^d I)$  \\
    BT-RNN backward & $\mathcal{O}(N d^2 I R^d J_{max})$         & $\mathcal{O}(R^d I)$  \\
    \hline
    \end{tabular}
    \end{center}
    \caption{Comparison of complexity and memory usage of vanilla RNN, Tensor-Train representation RNN (TT-RNN) \cite{novikov2015tensorizing, yang2017tensor} and our BT representation RNN (BT-RNN). In this table, the weight matrix's shape is $I \times J$. The input and hidden tensors' shapes are $I = I_1 \times \dots \times I_d$ and $J = J_1 \times \dots \times J_d$, respectively. Here, $J_{max} = \max_k(J_k), k \in [1, d]$. Both TT-RNN and BT-RNN are set in same rank $R$.}
    \label{tbl:complexity}
    \vspace{-2ex}
\end{table}

\vspace{-1ex}
\paragraph{Complexity Comparisons}
We analyze the time complexity and memory usage of RNN, Tensor Train RNN, and BT-RNN. The statistics are shown in Table \ref{tbl:complexity}. In our observation, both TT-RNN and BT-RNN hold lower computation complexity and memory usage than the vanilla RNN, since the extra hyper-parameters are several orders smaller than $I$ or $J$. As we claim that the suggested choice of Core-order is $d \in [2,5]$, the complexity of TT-RNN and BT-RNN should be comparable.







    \section{Experiments}

RNN is a versatile and powerful modeling tool widely used in various computer vision tasks. We design three challenging computer vision tasks-Action Recognition in Videos, Image Caption and Image Generation-to quantitatively and qualitatively evaluate proposed BT-LSTM against baseline LSTM and other low-rank variants such as Tensor Train LSTM (TT-LSTM). Finally, we design a control experiment to elucidate the effects of different hyper-parameters.



\subsection{Implementations}
Since operations in $\mathbf{f}_t$, $\mathbf{i}_t$, $\mathbf{o}_t$ and $\tilde{\mathbf{c}}{_t}$ follow the same computation pattern, we merge them together by concatenating $\mathbf{W}_f$, $\mathbf{W}_i$, $\mathbf{W}_o$ and $\mathbf{W}_c$ into one giant $\mathbf{W}$, and so does $\mathbf{U}$. This observation leads to the following simplified LSTM formulations: 
\begin{align}
    &\left({\mathbf{f}_t}', {\mathbf{i}_t}', {\tilde{\mathbf{c}}_t}', {\mathbf{o}_t}'\right) = \mathbf{W} \cdot \mathbf{x}_t + \mathbf{U} \cdot \mathbf{h}_{t-1} + \mathbf{b} \label{eqt:concat_lstm}, \\
   &\left(\mathbf{f}_t, \mathbf{i}_t, \tilde{\mathbf{c}}_t, \mathbf{o}_t）\right) = \left(\sigma({\mathbf{f}_t}'), \sigma({\mathbf{i}_t}'), \tanh({\tilde{\mathbf{c}}_t}'), \sigma({\mathbf{o}_t}')\right).
\end{align}
We implemented BT-LSTM on the top of simplified LSTM formulation with Keras and TensorFlow. The initialization of baseline LSTM models use the default settings in Keras and TensorFlow, while we use Adam optimizer with the same learning rate (lr) across different tasks.

\subsection{Quantitative Evaluations of BT-LSTM on the Task of Action Recognition in Videos}

\begin{figure}[t]
    \centering
    \subfigure[][Training loss of baseline LSTM, TT-LSTM and BT-LSTM.]{
    \includegraphics[width=0.45\textwidth]{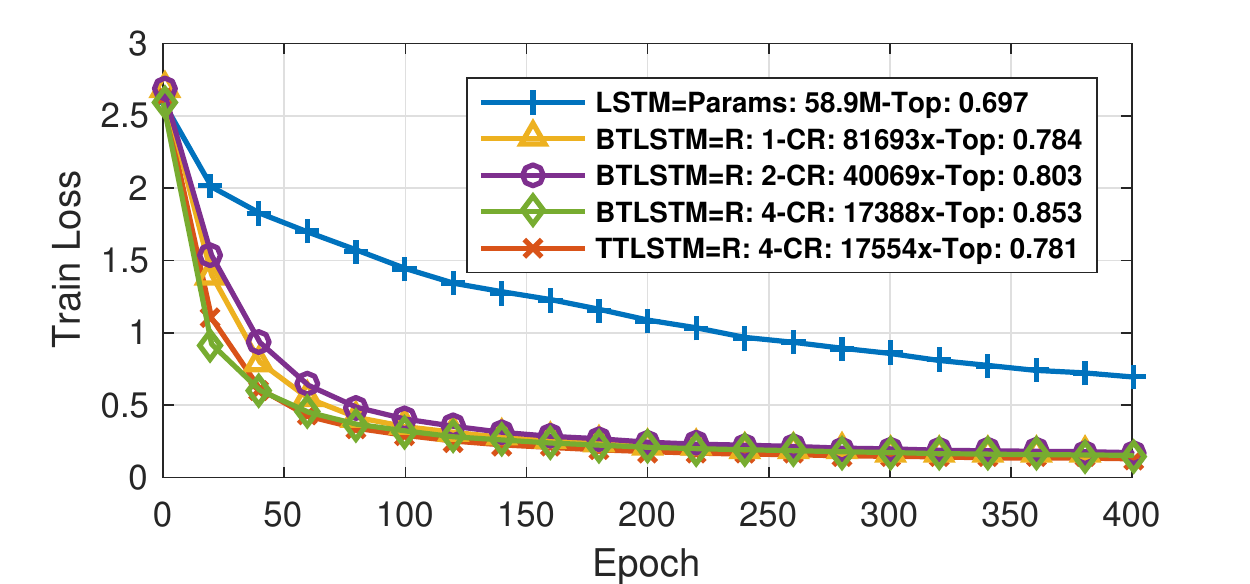} \label{fig:ucf11_loss}}

    \subfigure[][Validation Accuracy of baseline LSTM, TT-LSTM and BT-LSTM.]{
    \includegraphics[width=0.45\textwidth]{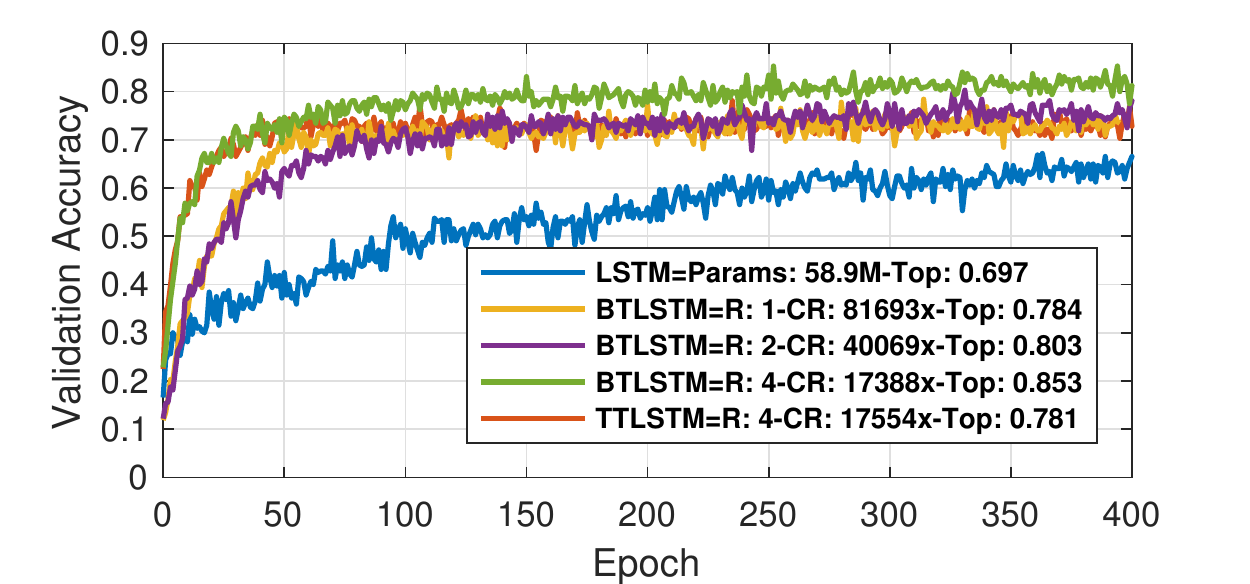} \label{fig:ucf11_acc}}
    \vspace{-1ex}
    \caption{Performance of different RNN models on the Action Recognition task trained with UCF11. CR stands for Compression Ratio; R is Tucker-rank, and Top is the highest validation accuracy observed in the training. Though BT-LSTM utilizes 17388 times less parameters than the vanilla LSTM (58.9 Millons), BT-LSTM demonstrates a 15.6\% higher accuracy improvement than LSTM. BT-LSTM also demonstrates and extra 7.2\% improvement over the TT-LSTM with comparable parameters.  }
    \vspace{-2ex}
    \label{fig:loss_convergence_ucf11}
\end{figure}

\begin{table}[t]
\small
\begin{center}
\begin{tabular}{c|c|c}
\hline
 & Method & Accuracy \\
\hline\hline
\multirow{3}{*}{\shortstack{Orthogonal \\ Approaches}} & Original \cite{liu2009recognizing} & 0.712 \\
 & Spatial-temporal \cite{liu2013spatial} & 0.761 \\
 & Visual Attention \cite{sharma2015action} & 0.850 \\
\hline
\multirow{3}{*}{\shortstack{RNN \\ Approaches}}
 & LSTM & {0.697} \\
 & TT-LSTM \cite{yang2017tensor} & 0.796 \\
 & BT-LSTM & \textbf{0.853} \\
\hline
\end{tabular}
\end{center}
\caption{State-of-the-art results on UCF11 dataset reported in literature, in comparison with our best model.}
\vspace{-2ex}
\label{tbl:ucf11_acc_state_of_the_art}
\end{table}


We use UCF11 YouTube Action dataset \cite{liu2009recognizing} for action recognition in videos. The dataset contains 1600 video clips, falling into 11 action categories. Each category contains 25 video groups, within each contains at least 4 clips. All video clips are converted to 29.97fps MPG\footnote{\url{http://crcv.ucf.edu/data/UCF11_updated_mpg.rar}}. We scale down original frames from $320 \times 240 \times 3$ to $160 \times 120 \times 3$, then we sample 6 random frames in ascending order from each video clip as the input data. For more details on the preprocessing, please refer to \cite{yang2017tensor}.

We use a single LSTM cell as the model architecture to evaluate BT-LSTM against LSTM and TT-LSTM in Fig. \ref{fig:loss_convergence_ucf11}. Please note there are other orthogonal approaches aiming at improving the model such as visual attention \cite{sharma2015action} and spatial-temporal \cite{liu2013spatial}. Since our discussion is limited to a single LSTM cell, we can always replace the LSTM cells in those high-level models with BT-LSTM to acquire better accuracies.
We set the hyper-parameters of BT-LSTM and TT-LSTM as follows: the factor tensor counts is $d = 4$; the shape of input tensor is $I_1 = 8, I_2 = 20, I_3 = 20, I_4 = 18$; and the hidden shape is $J_1 = J_2 = J_3 = J_4 = 4$; the rank of TT-LSTM is $R_1 = R_5 = 1, R_2 = R_3 = R_4 = 4$, while BT-LSTM is set to various Tucker-ranks.

Fig. \ref{fig:loss_convergence_ucf11} demonstrates the training loss and validation accuracy of BT-LSTM against LSTM and TT-LSTM under different settings. Table \ref{tbl:ucf11_acc_state_of_the_art} demonstrates the top accuracies of different models. From these experiments, we claim that:

1) \textbf{\textit{$\mathbf{\mathit{8\times 10^4}}$ times parameter reductions}}: The vanilla LSTM has 58.9 millons parameters in $\mathbf{W}$, while BT-LSTM deliveries better accuracies even with several orders of less parameters. The total parameters in BT-LSTM follows Eq. (\ref{eqt:number_of_parameters}). At Tucker-rank 1, 2, 4, BT-LSTM uses 721, 1470, and 3387 parameters, demonstrating compression ratios of 81693x, 40069x and 17388x, respectively.

2) \textbf{\textit{faster convergence}}: BT-LSTM demonstrates significant convergence improvement over the vanilla LSTM based on training losses and validation accuracies in Fig. \ref{fig:ucf11_loss} and Fig. \ref{fig:ucf11_acc}. In terms of validation accuracies, BT-LSTM reaches 60\% accuracies at epoch-16 while LSTM takes 230 epochs. The data demonstrates 14x convergence speedup. It is widely acknowledged that the model with few parameters is easier to train. Therefore, the convergence speedup majorly results from the drastic parameter reductions. At nearly same parameters, the training loss of BT-LSTM-4 also decreases faster than TTLSTM-4 ( epoches[0, 50] ), substantiating that BT model captures better spatial information than the Tensor Train model.

3) \textbf{\textit{better model efficiency}}: Though several orders of parameter reductions, BT-LSTM demonstrates extra 15.6\%  accuracies than LSTM. In addition, BT-LSTM also demonstrates  extra 7.2\% accuraies than TT-LSTM with comparable parameters.
In different Tucker-ranks, BT-LSTM converges to identical losses; but increasing Tucker ranks also improves the accuracy. This is consistent with the intuition since the high rank models capture additional relevant information.


\begin{figure}
    \centering
    \subfigure[][LSTM, \#Params:1.8M]{\includegraphics[width=0.45\columnwidth]{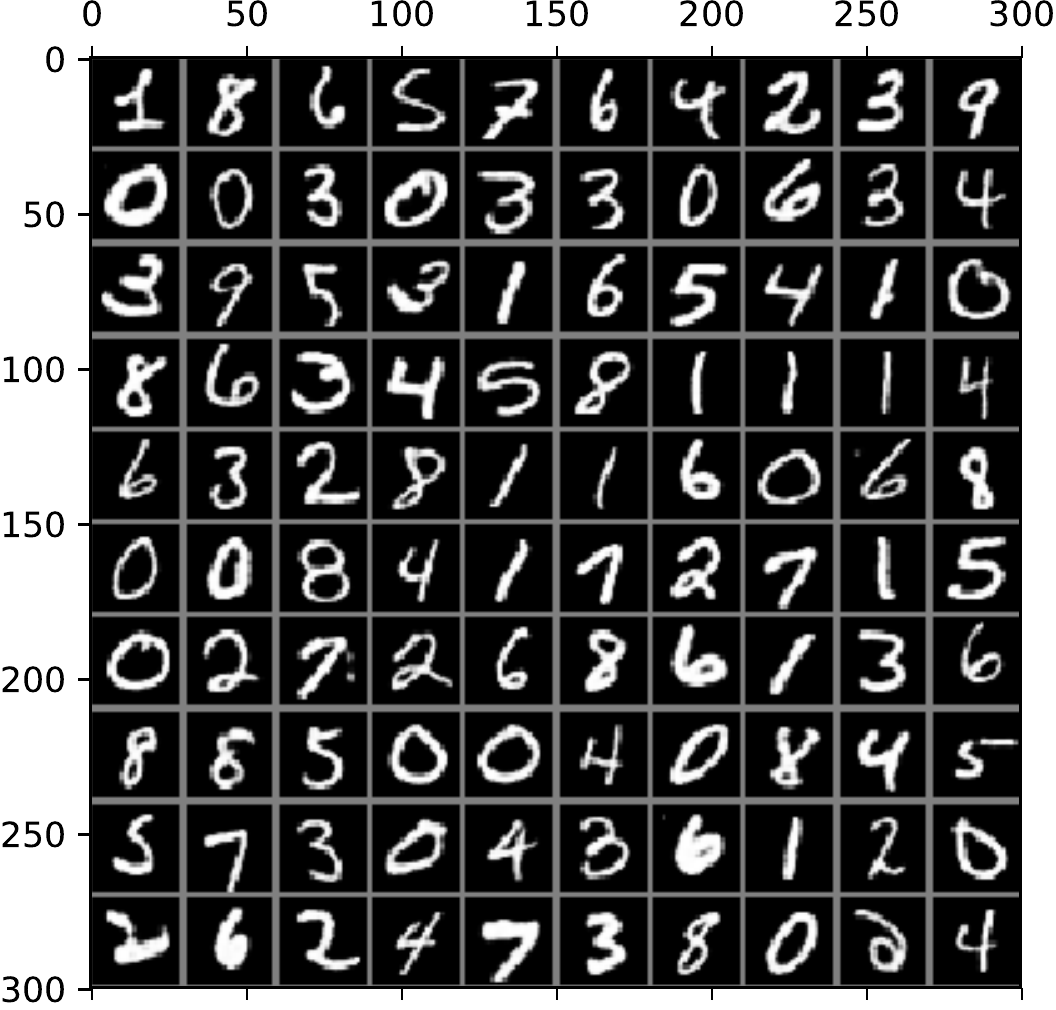}} \quad 
    \subfigure[][BT-LSTM, \#Params:1184]{\includegraphics[width=0.45\columnwidth]{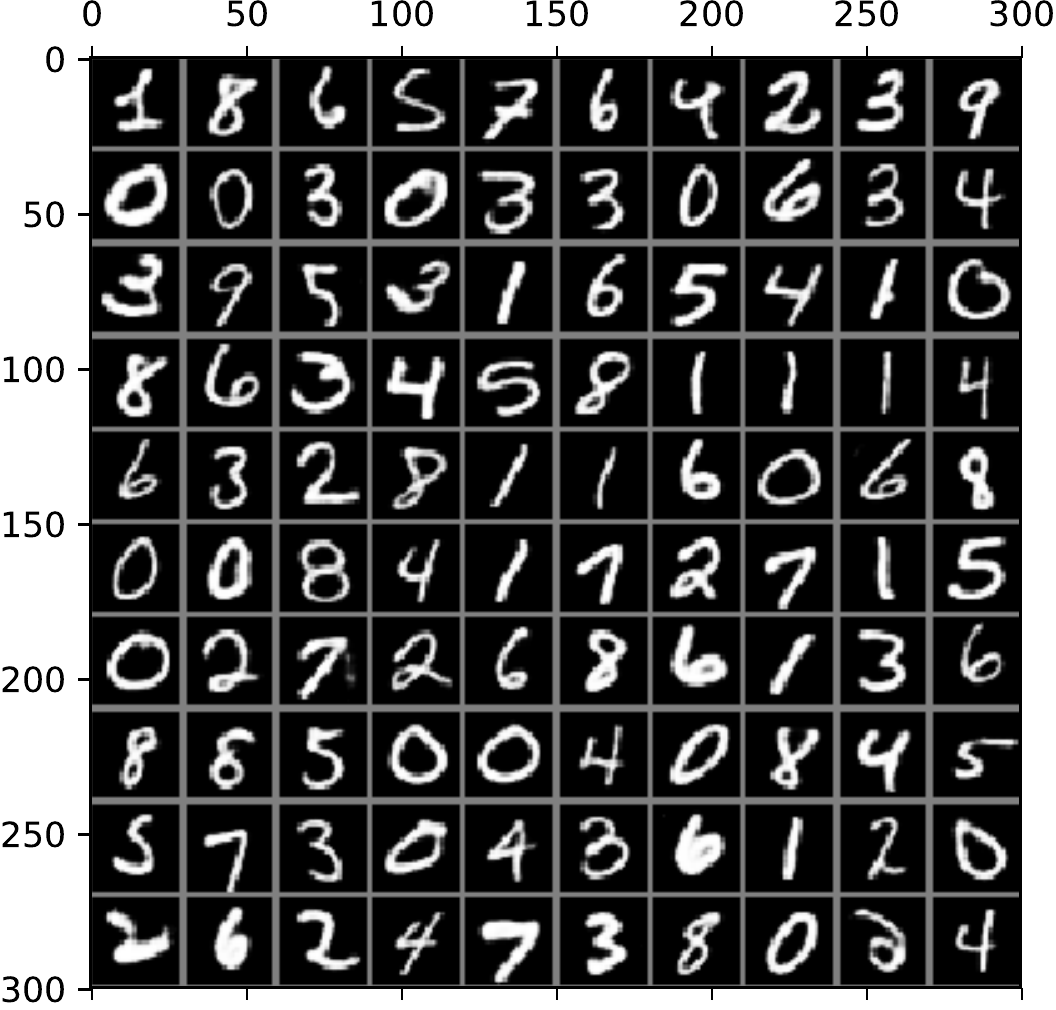}}
    \vspace{-1ex}
    \caption{Image Generation: generating MNIST style digits with LSTM and BT-LSTM based model. The results are merely identical, while the parameters of BT-LSTM is 1577 times less.}
    \vspace{-2ex}
    \label{fig:draw_all}
\end{figure}

\begin{figure*}[t]
    \captionsetup[subfigure]{labelformat=empty}
    \centering
    \subfigure[][ \scriptsize{ 
                  \textbf{LSTM}: A train traveling down tracks next to a forest. \newline
                  \textbf{TT-LSTM}: A train traveling down train tracks next to a forest. \newline
                  \textbf{BT-LSTM}: A train traveling through a lush green forest. }]{
    \includegraphics[width=0.20\textwidth, height=0.25\columnwidth]{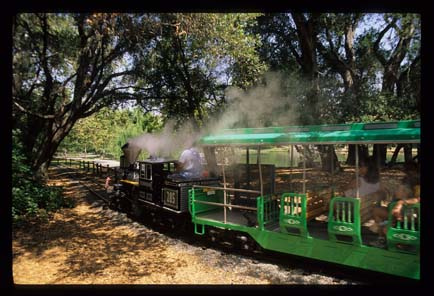}} \quad
    \subfigure[][\scriptsize{ \textbf{LSTM}: A group of people standing next to each other. \newline
                  \textbf{TT-LSTM}: A group of men standing next to each other. \newline
                  \textbf{BT-LSTM}: A group of people posing for a photo.
                }]{
        \includegraphics[width=0.20\textwidth, height=0.25\columnwidth]{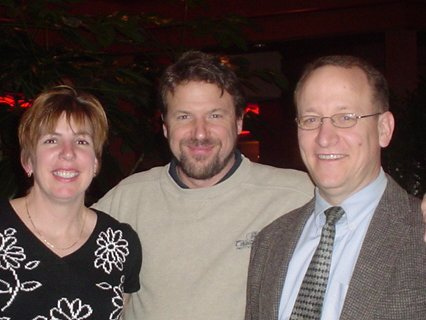}} \quad
    \subfigure[][\scriptsize{ \textbf{LSTM}: A man and a dog are standing in the snow. \newline 
                  \textbf{TT-LSTM}: A man and a dog are in the snow. \newline
                  \textbf{BT-LSTM}: A man and a dog playing with a frisbee.
                }]{
        \includegraphics[width=0.20\textwidth, height=0.25\columnwidth]{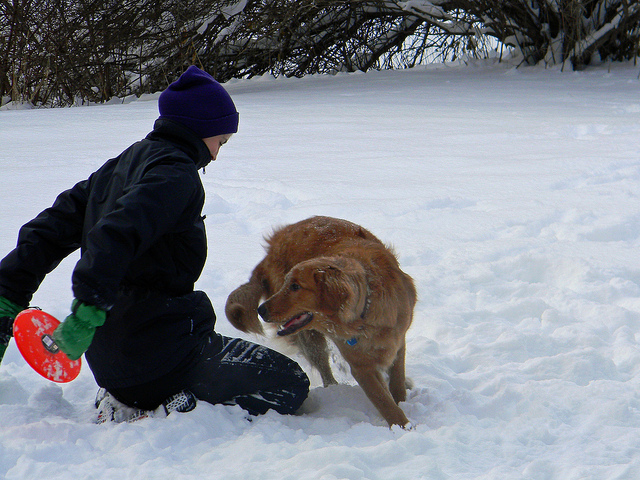}} \quad
    \subfigure[][\scriptsize{ \textbf{LSTM}: A large elephant standing next to a baby elephant. \newline 
                  \textbf{TT-LSTM}: An elephant walking down a dirt road near trees. \newline
                  \textbf{BT-LSTM}: A large elephant walking down a road with cars.
                }]{
		\includegraphics[width=0.20\textwidth, height=0.25\columnwidth]{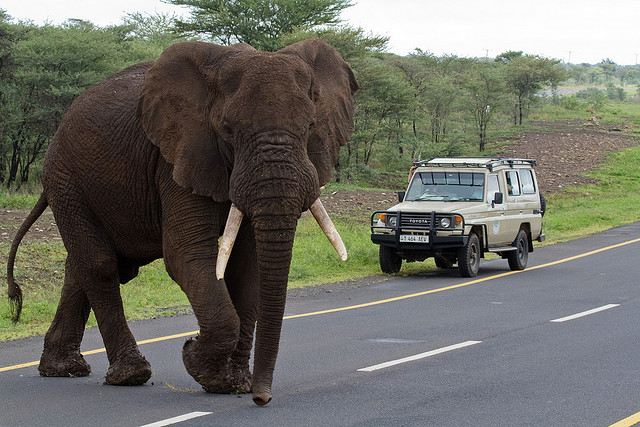}} \quad
    \caption{Results of image caption in MSCOCO dataset.}
    \label{fig:mscoco}
\end{figure*}


\subsection{Qualitative Evaluations of BT-LSTM on Tasks of Image Generation and Image Captioning}
We also conduct experiments on Image Generation and Image Captioning to further substantiate the effciency of BT-LSTM.

\vspace{-2ex}
\paragraph{Task 1: Image Generation}
Image generation intends to learn latent representation from images, then it tires to generate new image of same style from the learned model. The model for this task is Deep Recurrent Attentive Writer (DRAW) \cite{gregor2015draw}. It uses an encoder RNN network to encode images into latent representations; then an decoder RNN network decodes the latent representations to construct an image. we substitute LSTM in encoder network with our BT-LSTM. 

In this task, encoder network must capture sufficient correlations and visual features from raw images to generate high quality of feature vectors. As shown in Fig. \ref{fig:draw_all}, both LSTM and BT-LSTM model generate comparable images.

\vspace{-2ex}
\paragraph{Task 2: Image Captioning}
Image Captioning intends to describe the content of an image. We use the model in Neural Image Caption\cite{vinyals2015show} to evaluate the performance of BT-LSTM by replacing the LSTM cells.

The training dataset is MSCOCO \cite{lin2014microsoft}, a large-scale dataset for the object detection, segmentation, and captioning. Each image is scaled to $224\times 224$ in RGB channels and subtract the channel means as the input to a pretrained Inception-v3 model.

Fig. \ref{fig:mscoco} demonstrates the image captions generated by BT-LSTM and LSTM. It is obvious that both BT-LSTM, TT-LSTM and LSTM can generate proper sentences to describe the content of an image, but with little improvement in BT-LSTM. Since the input data of BT model is a compact feature vector merged with the embedding images features from Inception-v3 and language features from a word embedding network, our model demonstrates the qualitative improvement in captioning. The results also demonstrate that BT-LSTM captures local correlations missed by traditional LSTM.

\begin{figure}[t]
    \centering 
	\captionsetup[sub]{font=scriptsize,labelfont={bf,sf}}

    \subfigure[][\small Truth:$\mathbf{W}^{\prime}$ P=4096]{\includegraphics[height=0.2\columnwidth]{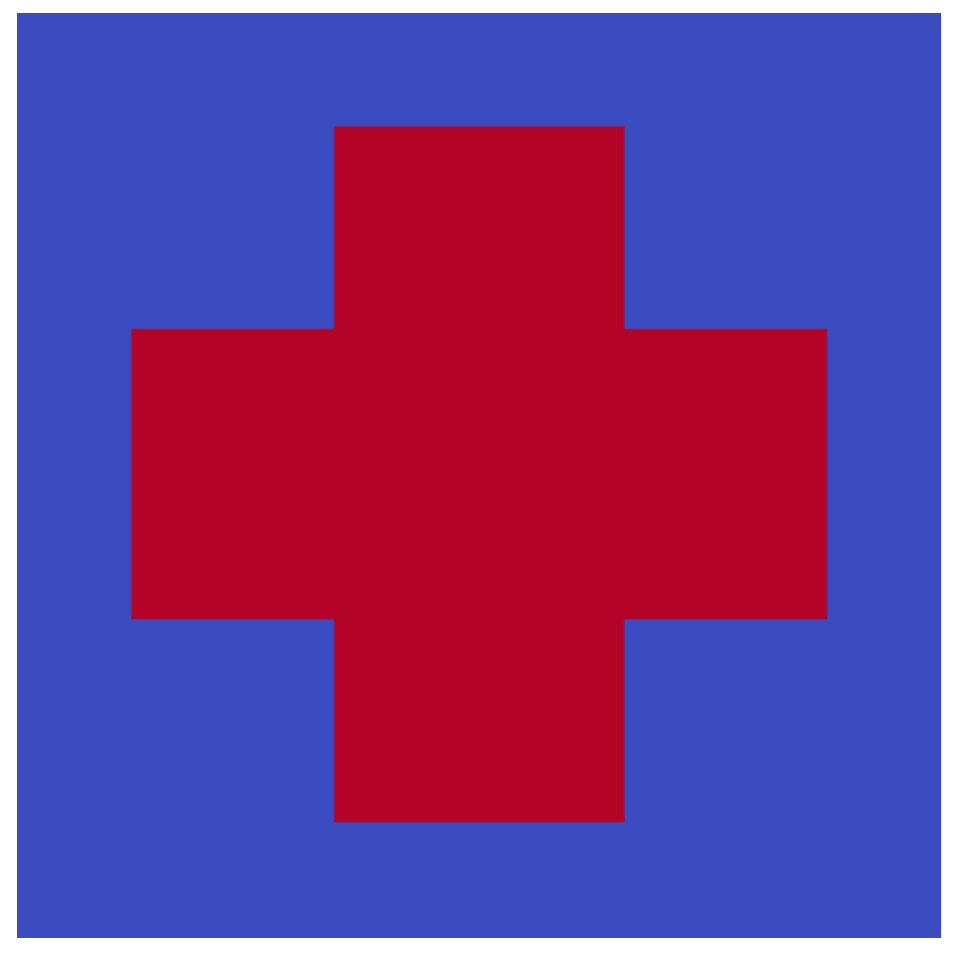} \label{fig:synthetic_org}} \quad 
    \subfigure[][$\mathbf{y}=\mathbf{W}\cdot\mathbf{x}$, P=4096]{\includegraphics[height=0.2\columnwidth]{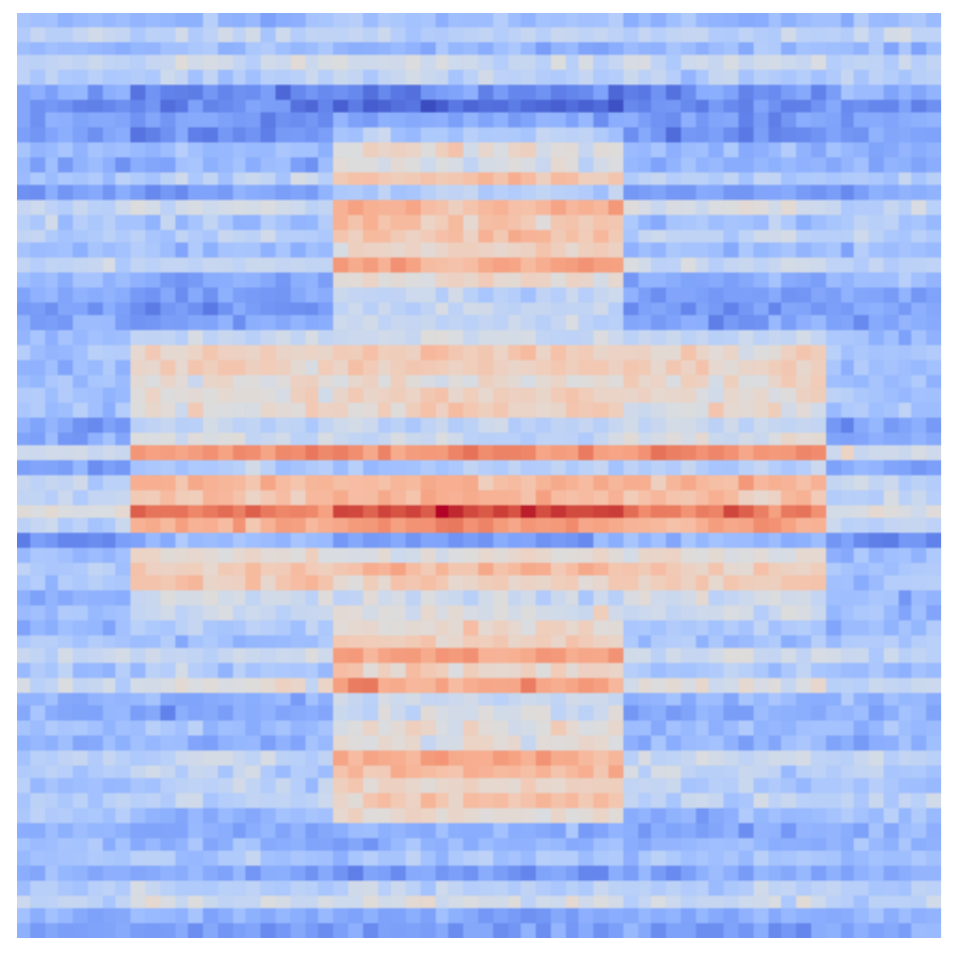} \label{fig:synthetic_linear}} \quad 
    \subfigure[][d=2, R=1, N=1, P=129]{\includegraphics[height=0.2\columnwidth]{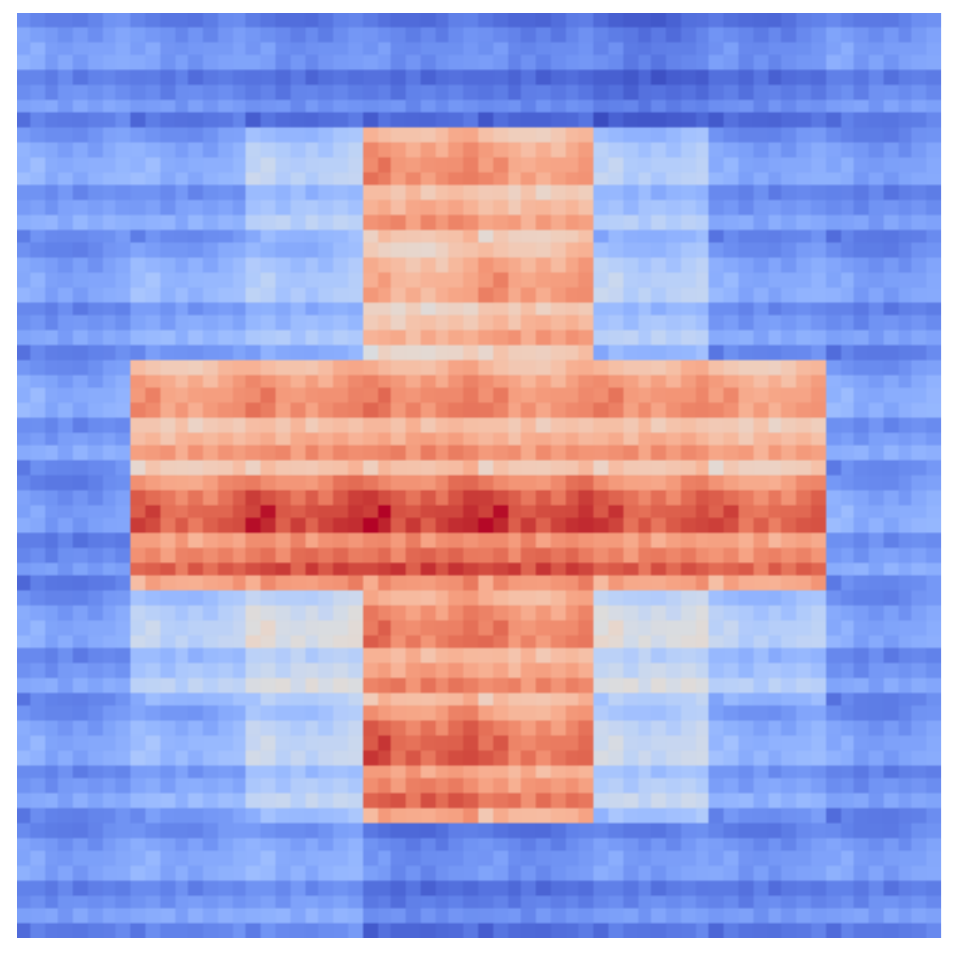} \label{fig:synthetic_d2r1n1}}  \quad 
	\\
	
    \subfigure[][d=2, R=4, N=1, P=528]{\includegraphics[height=0.2\columnwidth]{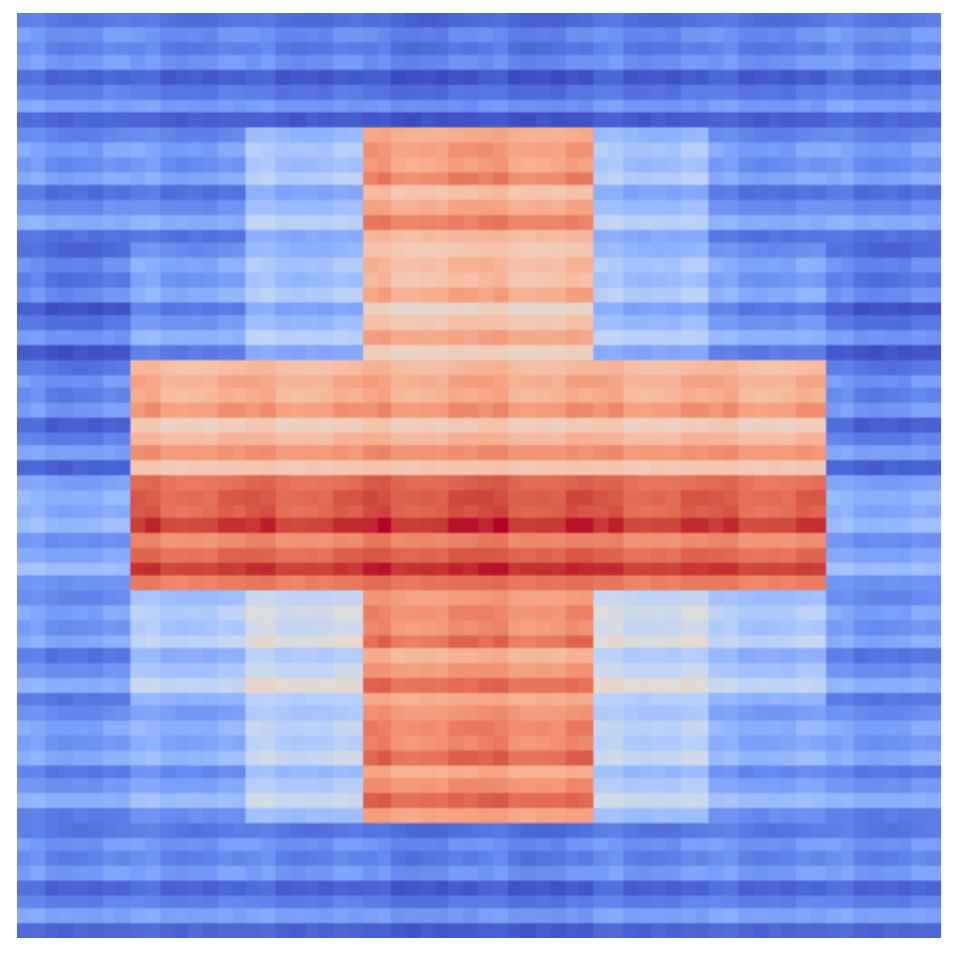} \label{fig:synthetic_d2r4n1}} \quad
    \subfigure[][d=2, R=1, N=2, P=258]{\includegraphics[height=0.2\columnwidth]{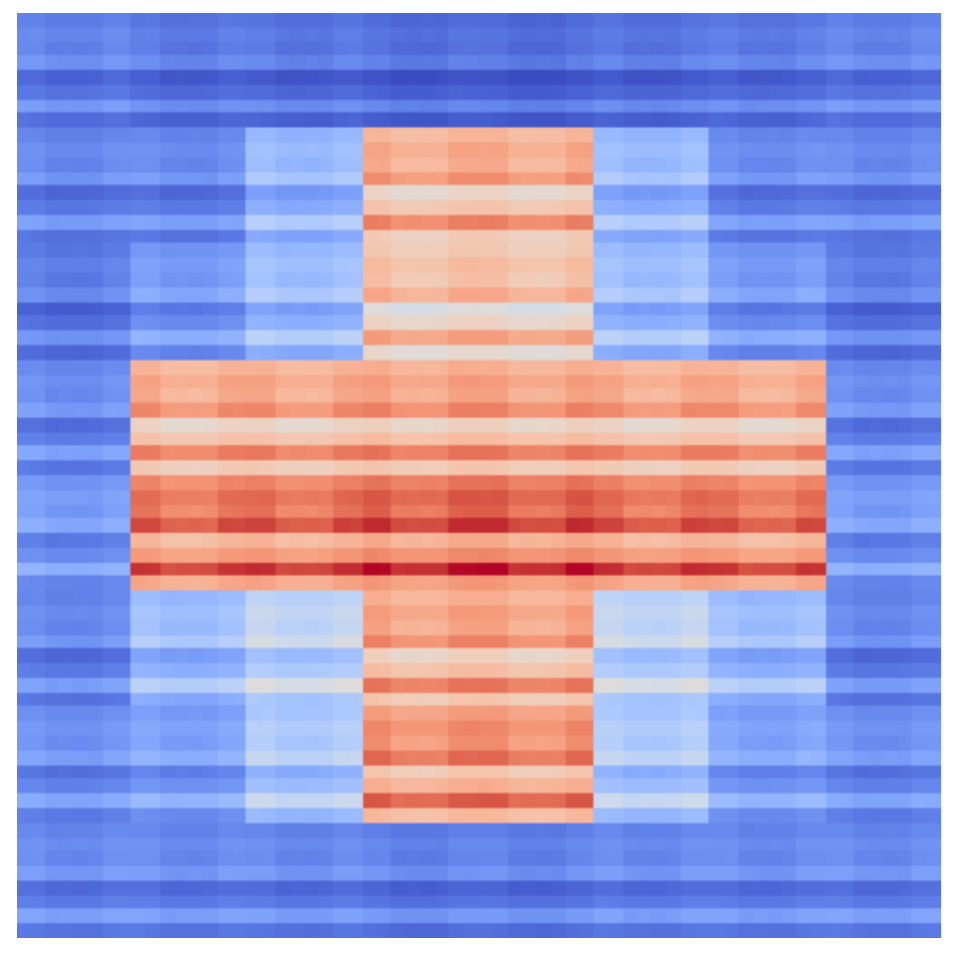} \label{fig:synthetic_d2r1n2}}  \quad
    \subfigure[][d=4, R=4, N=1, P=384]{\includegraphics[height=0.2\columnwidth]{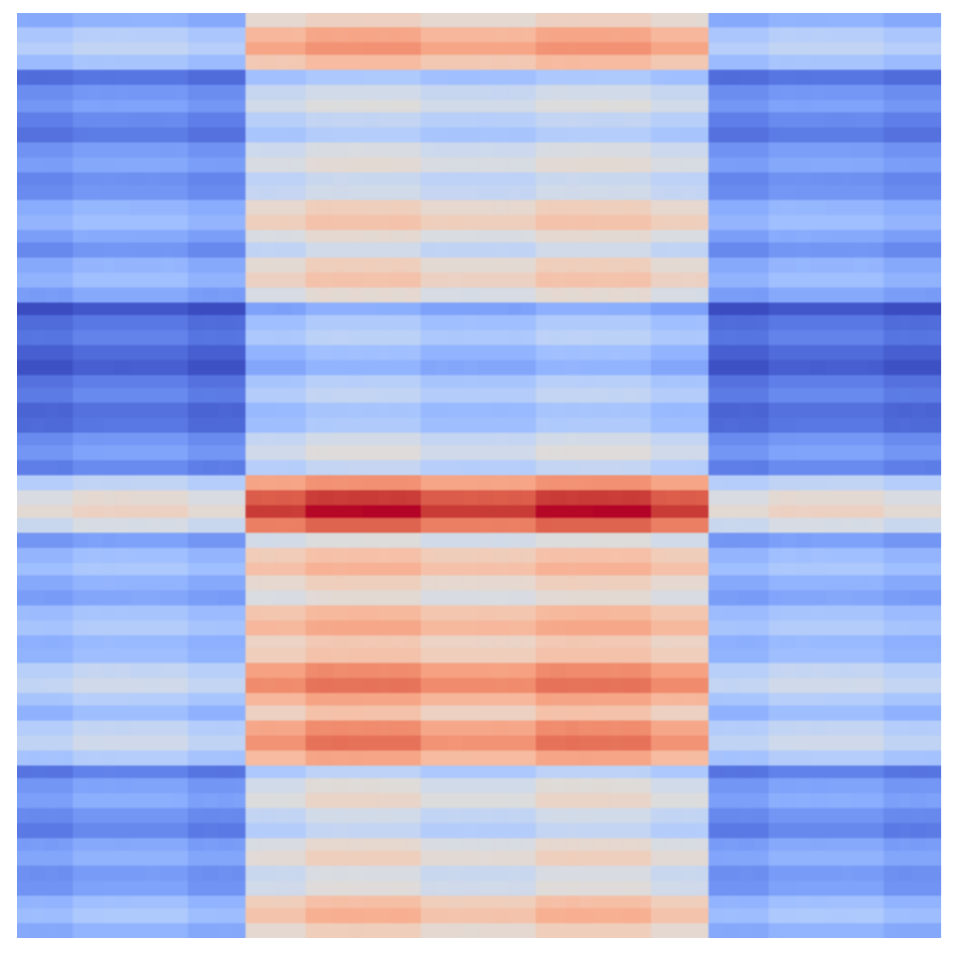} \label{fig:synthetic_d4r4n1}} \quad

    \vspace{-1ex}
    \caption{The trained $W$ for different BT-LSTM settings. The closer to (a), the better $\mathbf{W}$ is. }
    \vspace{-2ex}
    \label{fig:synthetic_all}
\end{figure}


\subsection{Sensitivity Analysis on Hyper-Parameters}
There are 3 key hyper-parameters in BT-LSTM, which are core-order $d$, Tucker-rank $R$ and CP-rank $N$. In order to scrutinize the impacts of these hyper-parameters, we design a control experiment illustrate their effects.

We try to sample y from the distribution of $\mathbf{y} = \mathbf{W}^{\prime} \cdot \mathbf{x}$, where $\mathbf{x},\mathbf{y} \in \mathbb{R}^{64}$. Each $\mathbf{x}$ is generated from a Gaussian distribution $\mathcal{N}(0, 0.5)$. We also add a small noise into $\mathbf{x}$ to avoid overfitting. $\mathbf{y}$ is generated by plugging $\mathbf{x}$ back to $\mathbf{y} = \mathbf{W}^{\prime} \cdot \mathbf{x}$. Given $\mathbf{x}$ and $\mathbf{y}$, we randomly initlize $\mathbf{W}$, and start training. Eventually, $\mathbf{W}$ should be similar to $\mathbf{W^{\prime}}$ since $\mathbf{x}$ and $\mathbf{y}$ drawn from the distribution of $\mathbf{y} = \mathbf{W}^{\prime} \cdot \mathbf{x}$. Please note that the purpose of this experiment is to evaluate the impact of the BT model on different parameter settings, despite these are many other good methods such as L1 regularization and Lasso regularization, to recover the $\mathbf{W}$ weight matrix.

\textbf{Core-order ($d$): }
Parameters goes down if $d$ grows and $d < 5$. Parameters reduce about 1.3 times from Fig. \ref{fig:synthetic_d2r4n1} to Fig. \ref{fig:synthetic_d4r4n1}; and $d$ increase from 2 to 4. With less parameters, the reconstructed $\mathbf{W}$ deteriorates quickly. We claim that high Core-order $d$ loses important spatial information, as tensor becomes too small to capture enough latent correlations. This result is consistent with our declaration. 

\textbf{Tucker-rank ($R$): }
the rank $R$ take effectiveness exponentially to the parameters. By comparing Fig. \ref{fig:synthetic_d2r1n1} and Fig. \ref{fig:synthetic_d2r4n1}, When $R$ increases from 1 to 4, BT model has more parameters to capture sufficient information from input data, obtaining a more robust model. 

\textbf{CP-rank ($N$): }
CP-rank contributes to the number of parameters linearly, playing an important role when $R$ is small. By comparing Fig. \ref{fig:synthetic_d2r1n1} and Fig. \ref{fig:synthetic_d2r1n2}, we can see that the latter result has less noise in figure, showing that a proper CP-rank setting will lead to a more robust model, since we use multiple Tucker models to capture information from input data.

%
%

    \section{Conclusion}
We proposed a Block-Term RNN architecture to address the redundancy problem in RNNs. By using a Block Term tensor decomposition to prune connections in the input-to-hidden weight matrix of RNNs, we provide a new RNN model with a less number of parameters and stronger correlation modeling between feature dimensions, leading to easy model training and improved performance. Experiment results on a video action recognition data set show that our BT-RNN architecture can not only consume several orders fewer parameters but also improve the model performance over standard traditional LSTM and the TT-LSTM.
The next works are to 1) explore the sparsity in factor tensors and core tensors of BT model, further reducing the number of model parameters; 2) concatenate hidden states and input data for a period of time, respectively, extracting the temporal features via tensor methods; 3) quantify factor tensors and core tensors to reduce memory usage.


    
    \section*{Acknowledgment}
    This paper was in part supported by a grant from the Natural Science Foundation of China (No.61572111),  1000-Talent Program Startup Funding (A1098531023601041,G05QNQR004) and a Fundamental Research Fund for the Central Universities of China (No. A03017023701). Zenglin Xu is the major corresponding author.

    {\small
    \bibliographystyle{ieee}
    \bibliography{egbib}
    }

\end{document}